%% file: main.tex
\documentclass{IEEEoj-data}
\usepackage{cite}
\usepackage{amsmath,amssymb,amsfonts}
\usepackage{graphicx,color}
\usepackage{xcolor}
\usepackage{textcomp}
\usepackage{hyperref}
\usepackage{url}
\usepackage[english = american]{csquotes} % Makes "" work properly
\usepackage{url,multirow,comment,hyperref,amsmath,graphicx,amsmath,subfigure,listings,tabularx,psfrag,wrapfig,cite,verbatim,array, algorithm, algorithmicx, algpseudocode}
\usepackage[
locale = DE % comma as decimal mark
]{siunitx} %riad added this
\usepackage{caption} 
\usepackage{xcolor,mathtools}
\usepackage{setspace} %RWR
\usepackage{multirow}

\usepackage[colorinlistoftodos]{todonotes}
\presetkeys{todonotes}{fancyline, color=blue!30,size=\tiny}{}

\def\BibTeX{{\rm B\kern-.05em{\sc i\kern-.025em b}\kern-.08em
    T\kern-.1667em\lower.7ex\hbox{E}\kern-.125emX}}
\AtBeginDocument{\definecolor{ojcolor}{cmyk}{0.93,0.59,0.15,0.02}}

\begin{document}
\receiveddate{13 March, 2025}
\reviseddate{11 April, 20xx}
\accepteddate{00 April, 20xx}
\publisheddate{00 May, 20xx}
\currentdate{24 June, 20xx}
\doiinfo{DD.2024.0624000}
%dhandeep changed according to review
%\title{{\color{black}\textbf{Meta:}} 
%\title{{\color{black}\textbf{Descriptor:}} 
\title{ 
\textit{Event-Based Crossing Dataset} (EBCD) }
%\title{Face Detection Dataset for Programmable Threshold-Based Event-Vision }

\author{
Joey~Mulé,
Dhandeep~Challagundla,
Rachit~Saini,~\IEEEmembership{Student Member,~IEEE},
and
Riadul~Islam,~\IEEEmembership{Senior Member,~IEEE}
%Chad~Howard,  
%and Ryan~Robucci,~\IEEEmembership{Member,~IEEE}

\thanks{J. Mulé, R. Islam, R. Kankipati, S. Jalapally, D. Challagundla, and R. Robucci are with the Department 
of Computer Science and Electrical Engineering, University of Maryland, Baltimore County, 
MD 21250, USA e-mail: {\{jmule2, riaduli, rohithk1, surajkj1, robucci\}@umbc.edu}.}
\thanks{C. Howard and C. Rizk are with the FRIS Inc. (d/b/a Oculi),
5520 Research Park Drive Catonsville MD, 21228, USA e-mail: {\{chad.howard, charbel.rizk\}@oculi.ai}}
\thanks{This work was supported in part by National Science Foundation (NSF) award number: 2138253, Maryland Industrial Partnerships (MIPS) program award number: MIPS0012, and the UMBC Startup grant.}
\thanks{Copyright (c) 2025 IEEE. Personal use of this material is permitted. 
However, permission to use this material for any other purposes must be 
obtained from the IEEE by sending an email to pubs-permissions@ieee.org.}
}

\markboth{IEEE Transactions on xxxxx}{Shell \MakeLowercase{\textit{et al.}}: ??????}

%\maketitle

\begin{abstract}

%\doublespacing %RWR

% IEEE Data Descriptions accepts articles with a recommended length of 4—8 pages (with a maximum of 10 pages for an initial submission and a maximum of 12 pages for revised submissions).

%**NOTE (NEED TO GO THROUGH PAPER, REVISE AND ADD CITATIONS WHERE NEEDED). ALSO NEED TO COMPLETE ACKNOWLEDGMENTS**

% Joey - Will revise
Event-based vision revolutionizes traditional image sensing by capturing asynchronous intensity variations rather than static frames, enabling ultrafast temporal resolution, sparse data encoding, and enhanced motion perception. While this paradigm offers significant advantages, conventional event-based datasets impose a fixed thresholding constraint to determine pixel activations, severely limiting adaptability to real-world environmental fluctuations. Lower thresholds retain finer details but introduce pervasive noise, whereas higher thresholds suppress extraneous activations at the expense of crucial object information. To mitigate these constraints, we introduce the Event-Based Crossing Dataset (EBCD), a comprehensive dataset tailored for pedestrian and vehicle detection in dynamic outdoor environments, incorporating a multi-thresholding framework to refine event representations. By capturing event-based images at ten distinct threshold levels (4, 8, 12, 16, 20, 30, 40, 50, 60, and 75), this dataset facilitates an extensive assessment of object detection performance under varying conditions of sparsity and noise suppression. We benchmark state-of-the-art detection architectures—including YOLOv4, YOLOv7, EfficientDet-b0, MobileNet-v1, and Histogram of Oriented Gradients (HOG)—to experiment upon the nuanced impact of threshold selection on detection performance. By offering a systematic approach to threshold variation, we foresee that EBCD fosters a more adaptive evaluation of event-based object detection, aligning diverse neuromorphic vision with real-world scene dynamics. We present the dataset as publicly available to propel further advancements in low-latency, high-fidelity neuromorphic imaging: \url{https://ieee-dataport.org/documents/event-based-crossing-dataset-ebcd}

{\textcolor{ieeedata}{\abstractheadfont\bfseries{IEEE SOCIETY/COUNCIL}}}     Signal Processing Society (SPS)\\ 
 %\\
 {\textcolor{ieeedata}{\abstractheadfont\bfseries{DATA DOI/PID}}}     10.21227/ahdq-g045\\ 
 {\textcolor{ieeedata}{\abstractheadfont\bfseries{DATA TYPE/LOCATION}}}   Images; Maryland, USA

\end{abstract}
\begin{IEEEkeywords}
Pedestrian detection, Vehicle detection, dynamic vision sensing (DVS), sparse vision, convolutional neural network (CNN), Pedestrian crossing dataset.
\end{IEEEkeywords}
\maketitle

%\doublespacing %RWR

\input{introduction.tex}
\input{existing_approaches.tex}
\input{proposed_method.tex}

\input{applied_analysis.tex}

\input{analysis.tex}

% dhandeep removed conclusion according to descriptor requirements
%\input{conclusion.tex}
\section*{Source Code and Scripts}
In this study, we utilized the NTU Pedestrian video dataset and conducted an analysis using open-source, commonly known SOTA models and architectures. Sample dataset analysis scripts and miscellaneous information is available at the following GitHub repository: \url{https://github.com/joeduman/Thresholded_event-based-crossing-dataset}.

% Need to complete
\section*{Acknowledgments}

Conceptualization, R. Islam and J. Mule; Dataset, original draft, and analysis, J. Mule, D. Challagundla, R. Saini, and R. Islam; Review and editing, R. Islam, J. Mule; funding acquisition, R. Islam. 

This research was funded by the National Science Foundation (NSF) under award number 2138253 and the UMBC Startup grant. Special thanks to Rohith Kankipati, Suraj Jalapally, Rachit Saini, Ryan Robucci of UMBC, and Chad Howard of Oculi for assisting in annotation, analysis, and discussion. 

The article authors have declared no conflicts of interest.

% J. Mulé, %conception, visualizations, dataset evaluation
% R. Islam, %annotations, P.I. supervision, motivation, algorithm development
% D. Challagundla, %annotations
% Special thanks to those who contributed to the dataset annotations and partial analysis:
% Suraj Jalapally, %annotations and dataset evalutaion
% Rohith Kankipati, %annotations and dataset evaluation
% Ryan Robucci, %

% Chad Howard, % dataset providence
% R. Saini % annotations

% FILL

% UP

% REST

\begin{table*}[h!]
	\caption{ All metrics are in percentages. YOLO-v4 and YOLO-v7 perform consistently well across thresholds, while EfficientDet-b0 excels in precision but struggles with AP at higher thresholds due to low recall. MobileNet-v1 peaks at $T_{16}$, but declines at higher thresholds. HOG performs best at $T_{75}$, but struggles overall due to localization challenges in tight bounding boxes.
		\label{tab:benchmarks}}
    \centering
    \vspace{-0.150cm}
    \begin{tabular}{|c|c|c|c|c|c|c|c|c|}
    \hline
        \multirow{2}{*}{ Models} &\multirow{2}{*}{\# of Parameters} & \multirow{2}{*}{$T_h$} &  \multicolumn{3}{c|}{Metrics} &  \multirow{2}{*}{IoU (\%)} & \multirow{2}{*}{AP50} & \multirow{2}{*}{AP75} \tabularnewline
			\cline{4-6}
        ~ & ~ & ~ & Precision & Recall & F1-score & ~ & ~ & ~  \\ \hline
        \multirow{10}{*}{YOLO-v4~\cite{Bochkovskiy_yolov4:2020}} &  \multirow{10}{*}{60.3M} & 4 & 96.00 & 82.00 & 88.00 & 78.74 & 96.51 & 71.06 \multirow{10}{*}{}  \tabularnewline 
			\cline{3-9}
        ~ & ~ & 8 & 96.00 & 82.00 & 88.00 & 79.08 & 97.20 & 75.04  \tabularnewline 
                \cline{3-9}
        ~ & ~ & 12 & 96.00 & 80.00 & 87.00 & 79.38 & 95.93 & 74.43 \tabularnewline 
                \cline{3-9}
        ~ & ~ & 16 & 94.00 & 81.00 & 87.00 & 78.06 & 93.57 & 76.80 \tabularnewline 
                \cline{3-9}
        ~ & ~ & 20 & 94.00 & 84.00 & 87.00 & 77.76 & 96.44 & 73.96 \tabularnewline 
                \cline{3-9}
        ~ & ~ & 30 & 93.00 & 79.00 & 85.00 & 75.95 & 93.90 & 67.64 \tabularnewline 
                \cline{3-9}
        ~ & ~ & 40 & 93.00 & 75.00 & 83.00 & 75.30 & 91.00 & 62.41 \tabularnewline 
                \cline{3-9}
        ~ & ~ & 50 & 89.00 & 73.00 & 80.00 & 68.45 & 87.25 & 43.20 \tabularnewline 
                \cline{3-9}
        ~ & ~ & 60 & 91.00 & 69.00 & 78.00 & 71.49 & 84.00 & 48.92 \tabularnewline 
                \cline{3-9}
        ~ & ~ & 75 & 93.00 & 63.00 & 75.00 & 70.71 & 77.94 & 35.77 \\ \hline
         \multirow{10}{*}{YOLO-v7~\cite{Wang_yolov7:2022}} &  \multirow{10}{*}{25.2M} & 4 & 98.00 & 78.00 & 87.00 & 81.87 & 93.11 & 75.77 \tabularnewline 
			\cline{3-9}
        ~ & ~ & 8 & 98.00 & 78.00 & 87.00 & 81.84 & 93.24 & 76.55  \tabularnewline 
                \cline{3-9}
        ~ & ~ & 12 & 97.00 & 78.00 & 87.00 & 81.71 & 91.45 & 76.32 \tabularnewline 
                \cline{3-9}
        ~ & ~ & 16 & 97.00 & 78.00 & 86.00 & 81.31 & 91.28 & 76.84 \tabularnewline 
                \cline{3-9}
        ~ & ~ & 20 & 97.00 & 79.00 & 87.00 & 81.82 & 95.31 & 78.06 \tabularnewline 
                \cline{3-9}
        ~ & ~ & 30 & 97.00 & 76.00 & 85.00 & 81.49 & 93.23 & 75.64 \tabularnewline 
                \cline{3-9}
        ~ & ~ & 40 & 95.00 & 72.00 & 82.00 & 79.15 & 88.62 & 68.18 \tabularnewline 
                \cline{3-9}
        ~ & ~ & 50 & 93.00 & 70.00 & 80.00 & 76.67 & 86.35 & 65.34 \tabularnewline 
                \cline{3-9}
        ~ & ~ & 60 & 95.00 & 69.00 & 80.00 & 76.63 & 83.31 & 57.18 \tabularnewline 
                \cline{3-9}
        ~ & ~ & 75 & 97.00 & 61.00 & 75.00 & 76.87 & 77.16 & 48.10 \\ \hline
                  \multirow{10}{*}{EfficientDet-b0~\cite{Tan_efficientdet:2020}} &  \multirow{10}{*}{3.9M} & 4 & 99.44 & 52.53 & 68.74 & 84.52 & 73.57 & 50.79  \tabularnewline 
			\cline{3-9}
        ~ & ~ & 8 & 97.88 & 54.83 & 70.29 & 84.94 & 74.33 & 51.94  \tabularnewline 
                \cline{3-9}
        ~ & ~ & 12 & 98.44 & 58.22 & 73.17 & 86.07 & 76.61 & 55.27 \tabularnewline 
                \cline{3-9}
        ~ & ~ & 16 & 99.21 & 57.03 & 72.43 & 86.33 & 76.22 & 55.38 \tabularnewline 
                \cline{3-9}
        ~ & ~ & 20 & 100.00 & 46.40 & 63.39 & 86.36 & 69.93 & 50.85 \tabularnewline 
                \cline{3-9}
        ~ & ~ & 30 & 99.39 & 48.59 & 65.27 & 85.54 & 71.08 & 50.75 \tabularnewline 
                \cline{3-9}
        ~ & ~ & 40 & 98.68 & 45.73 & 62.50 & 84.63 & 69.64 & 47.23 \tabularnewline 
                \cline{3-9}
        ~ & ~ & 50 & 98.84 & 38.23 & 55.14 & 85.00 & 64.07 & 42.00 \tabularnewline 
                \cline{3-9}
        ~ & ~ & 60 & 98.86 & 38.80 & 55.73 & 83.79 & 64.46 & 41.41 \tabularnewline 
                \cline{3-9}
        ~ & ~ & 75 & 99.19 & 37.07 & 53.96 & 82.96 & 63.07 & 40.31 \\ \hline
         
         \multirow{10}{*}{MobileNet-v1~\cite{Howard_mobilenets:2017}} &  \multirow{10}{*}{6.05M} & 4 & 67.70 & 6.50 & 11.90 & 74.68 & 0.00 & 0.00  \tabularnewline 
			\cline{3-9}
        ~ & ~ & 8 & 98.90 & 28.10 & 43.70 & 81.92 & 0.00 & 0.00  \tabularnewline 
                \cline{3-9}
        ~ & ~ & 12 & 98.30 & 26.20 & 41.40 & 82.32 & 26.16 & 18.88 \tabularnewline 
                \cline{3-9}
        ~ & ~ & 16 & 95.60 & 77.60 & 85.50 & 82.58 & 75.96 & 53.11 \tabularnewline 
                \cline{3-9}
        ~ & ~ & 20 & 94.10 & 36.80 & 52.90 & 81.46 & 35.58 & 25.18 \tabularnewline 
                \cline{3-9}
        ~ & ~ & 30 & 93.70 & 39.50 & 55.60 & 79.20 & 37.29 & 23.65 \tabularnewline 
                \cline{3-9}
        ~ & ~ & 40 & 93.40 & 37.60 & 53.50 & 77.59 & 35.46 & 19.42 \tabularnewline 
                \cline{3-9}
        ~ & ~ & 50 & 88.80 & 30.60 & 29.40 & 75.52 & 30.54 & 14.51 \tabularnewline 
                \cline{3-9}
        ~ & ~ & 60 & 85.40 & 32.20 & 46.90 & 73.36 & 24.65 & 8.23 \tabularnewline 
                \cline{3-9}
        ~ & ~ & 75 & 72.90 & 19.10 & 30.30 & 70.64 & 18.21 & 5.31 \\ \hline
        \multirow{10}{*}{HOG~\cite{HOG:2005}} &  \multirow{10}{*}{3780} & 4 & 66.67 & 1.42 & 2.79 & 10.60 & 0.00 & 0.00  \tabularnewline 
			\cline{3-9}
        ~ & ~ & 8 & 27.17 & 4.45 & 7.65 & 7.22 & 0.00 & 0.00  \tabularnewline 
                \cline{3-9}
        ~ & ~ & 12 & 13.16 & 3.56 & 5.60 & 10.39 & 0.00 & 0.00 \tabularnewline 
                \cline{3-9}
        ~ & ~ & 16 & 9.33 & 4.98 & 6.50 & 16.67 & 0.00 & 0.00 \tabularnewline 
                \cline{3-9}
        ~ & ~ & 20 & 4.37 & 3.20 & 3.70 & 17.74 & 0.00 & 0.00 \tabularnewline 
                \cline{3-9}
        ~ & ~ & 30 & 5.10 & 4.27 & 4.65 & 18.98 & 0.00 & 0.00 \tabularnewline 
                \cline{3-9}
        ~ & ~ & 40 & 9.07 & 8.54 & 8.80 & 24.08 & 0.00 & 0.00 \tabularnewline 
                \cline{3-9}
        ~ & ~ & 50 & 14.56 & 13.70 & 14.12 & 26.18 & 0.00 & 0.00 \tabularnewline 
                \cline{3-9}
        ~ & ~ & 60 & 13.67 & 12.46 & 13.04 & 25.87 & 0.00 & 0.00 \tabularnewline 
                \cline{3-9}
        ~ & ~ & 75 & 15.77 & 14.06 & 14.86 & 25.17 & 0.00 & 0.00 \\ \hline 
    \end{tabular}
    \vspace{-0.150cm}
\end{table*}
\clearpage

\bibliographystyle{IEEEtran}
\bibliography{main}

\end{document}

%% file: introduction.tex
\section*{Background}

%importance of implementing and architecting architectures for event-vision

% Algorithm development for smart sensors, .....

% Joey - maybe ok
The advancement of smart vision sensors that integrate both sensing and processing is critical for applications requiring low-latency, efficient perception. These sensors \cite{Lichtsteiner:2008, EBV:A_Suvey} operate differently from conventional imaging systems by capturing scene dynamics rather than transmitting full-frame data, reducing redundant information, and optimizing computational efficiency, similar to other spike-based computations~\cite{islam2023exploring, caporale2008spike, islam2024benchmarking}. Instead of capturing all raw data, event sensors perform online data filtering and selective in-plane feature computations for a given task \cite{Amir_dvs:2017, Gallego:2022}. This minimizes both downstream communication data flow and computational overhead within the vision system. However, this approach presents challenges in both algorithm development and data acquisition, particularly for outdoor environments where lighting, weather, and motion variability introduce additional complexities. 

Unlike standard frame-based cameras \cite{Lichtsteiner:2008}, event cameras, also known as neuromorphic or dynamic vision sensors, operate by asynchronously detecting changes in pixels over time \cite{Gallego:2022}, requiring specialized processing techniques to extract useful information for downstream vision tasks \cite{Rebecq:2019}. Due to the asynchronous nature of event sensors, they could naturally mask soft errors compared to existing radiation-tolerant systems~\cite{Islam_asicon:2011, mitra2007built, Islam_isqed:2012, islam2018low}.
A fundamental issue in event-driven vision is the lack of standardized models to emulate sensor behavior and generate realistic event datasets. In outdoor settings, this challenge is amplified due to unpredictable environmental conditions, such as fluctuating illumination, occlusions, and non-repetitive object movement. While frame-based vision systems rely on dense temporal sampling to capture motion \cite{Lichtsteiner:2008}, event imaging selectively encodes changes, often at a rate that does not permit traditional reconstruction techniques \cite{Rebecq:2019}. The adaptive pre-processing mechanisms of these sensors introduce further variability, making it difficult to establish uniform methodologies for analysis~\cite{Miao:2019}.

These difficulties extend to event-based pedestrian detection, where existing outdoor datasets \cite{1MEGAPIXEL, EventBasedPedestrain, EventBasedPedestrian_IEEE, Chen:2018, Boretti:2023} %Explore more about these datasets?
provide useful information and benchmarks. However, these datasets often employ fixed or narrow event-triggering thresholds, which can lead to excessive noise capture and may not effectively handle rapid illumination, shadows, and environmental changes such as wind, or falling rain. Additionally, the reliance on specific camera sensor configurations with static parameter tuning, restrics their adaptability to broader real-world pedestrian and vehicle tracking in scene understanding tasks.
To address these challenges, future event-based datasets must incorporate adaptive thresholding mechanisms to increase fidelity in temporal processing.

Ultimately, a multi-thresholding technique \cite{SEFD:2024} creates the foundation of the dataset to optimize event-based vision for low-bandwidth pedestrian detection applications. By offering images captured at a wider range of threshold ($T_h$) levels, such a dataset enables the calibration of neural network (NN) architectures to effectively handle sparse and varied visual inputs \cite{Afshar:2020}.
%Fill in space?

\subsection*{Motivation}
%Joey - ok
Event-based camera vision is a transformative approach to visual sensing that mimics how biological eyes process information, detecting changes in brightness at each pixel asynchronously. This results in significantly reduced non-informative data, lower latency, and improved efficiency, making them well-suited for real-time applications such as robotics, autonomous vehicles, and neuromorphic computing~\cite{Chakravarthi_survey:2024, islam2018negative}.
The critical advantages of event-based vision include its low power consumption, high temporal resolution, and robustness to challenging lighting conditions. These features enable superior motion detection, object tracking, and dynamic scene analysis, even in scenarios where conventional cameras struggle—such as high-speed environments or low-light settings. As event-based vision continues to evolve, it is poised to revolutionize fields such as biomedical imaging~\cite{Zhang_event:2024, Guo_eventlfm:2024}, augmented reality~\cite{Dong_sevar:2024}, and edge computing~\cite{Gokarn_poster:2024}, paving the way for more efficient and intelligent vision systems. 
Unlike other energy-efficient computation~\cite{Islam_dcmcs:2018, islam2011high, guthaus2017current, perez2024general, Islam_diff:2015, challagundla2022power, busia2024tiny}, event-based vision also holds immense potential for autonomous driving, where real-time perception and rapid response times are critical for safety and navigation. The high temporal resolution of event camera vision enables precise pedestrian and vehicle detection, particularly in dynamic urban environments where sudden movements and occlusions challenge traditional systems. The sparse pixel data representation allows for real-time processing on resource-constrained edge devices with accelerated runtime. 
%removed for space
These capabilities are essential for real-time object tracking and decision-making, enhancing the efficiency of intelligent transportation systems and next-generation autonomous systems.

\subsection*{Main Contributions}
%removed for space
%In this paper, we resolve the major shortcoming of available event-based vision-related benchmark datasets~\cite{Orchard:2015, Serrano-gotarredona_linares-barranco:2015}, where the data was collected considering a single threshold. The proposed tool flow will allow new neural architecture exploration and adaptive image sensing. In particular, 

In this research, we aim to shed light on the shortcoming of available event-based vision-related benchmark datasets~\cite{Orchard:2015, Serrano-gotarredona_linares-barranco:2015} that consider only a fixed threshold of data. In particular:
%Joey
\begin{itemize} \renewcommand{\labelitemi}{$\bullet$}
		
	   \item We present a new, innovative event based crossing dataset (EBCD), comprised of multiple programmable digital thresholds to decouple the challenges of modeling smart sensors and initial algorithm development.
       \item We detect multiple objects, including pedestrains, as well as passing vehicles.
          \item We analyzed pixel activity concerning extensive $T_h$ values to characterize event-based object detection.
	   \item We validate the effectiveness of the proposed dataset through training of industry-standard, state-of-the-art (SOTA) object detection and localization models.
        \item We evaluate with low to high $T_h$ values to distinguish a difference between pixel activity of indoor, controlled environments, to noisy outdoor scenarios with high constrast.
\end{itemize}

\subsection*{Existing Neural Network Architectures for Object Detection and Event-Based Vision}

Object detection and localization remain fundamental challenges in computer vision, particularly in dynamic environments where real-time processing and adaptability are crucial. Traditional frame-based approaches rely on high-resolution images, whereas event-based vision systems introduce a different paradigm, capturing asynchronous changes in the scene rather than static frames. The application of neural networks in object detection must therefore accommodate both conventional image data and the unique characteristics of event-driven inputs.

Among the most widely adopted object detection models is You Only Look Once (YOLO), a family of architectures designed for high-speed, single-shot detection~\cite{Redmon_yolo:2016}. The original YOLO model introduced a grid-based detection framework, dividing an image into an $S \times S$ grid and assigning bounding boxes based on object centers. This approach allows real-time inference but initially struggled with small and overlapping objects. YOLO-v4~\cite{Bochkovskiy_yolov4:2020} improved detection accuracy and efficiency by incorporating weighted residual connections (WRC) and cross-stage partial connections (CSPs), optimizing feature extraction while reducing computational overhead. YOLO-v7~\cite{Wang_yolov7:2022} further refined these capabilities through the Extended Efficient Layer Aggregation Network (E-ELAN), which enhances multi-scale feature fusion, making it better suited for handling sparse or rapidly changing inputs—such as those encountered in event-based vision.

Beyond YOLO-based models, alternative architectures have been developed to balance computational efficiency and detection accuracy. EfficientDet~\cite{Tan_efficientdet:2020}, developed by Google Brain, introduces a bi-directional feature pyramid network (BiFPN) and a compound scaling approach, optimizing model depth, resolution, and width for improved performance. The base variant, EfficientDet-b0, provides an efficient backbone for object detection tasks, particularly in resource-constrained environments where low-latency processing is required—aligning well with the demands of event-driven data processing.

For real-time detection on edge devices and embedded systems, MobileNet-v1~\cite{Howard_mobilenets:2017} was designed as an efficient convolutional neural network (CNN) optimized for mobile and low-power applications. Unlike conventional CNN architectures, MobileNet-v1 employs depthwise separable convolutions, a factorized approach that decomposes standard convolutions into separate depthwise and pointwise operations. MobileNet-v1 utilizes a width multiplier and a resolution multiplier to allow flexibility in balancing accuracy and efficiency.
While deep learning architectures have demonstrated remarkable success in object detection, classical methods such as the Histogram of Oriented Gradients (HOG) remain relevant, particularly in structured environments~\cite{HOG:2005}. HOG operates by computing gradient orientation histograms over localized regions, effectively capturing shape-based features. Though deep networks generally outperform traditional feature-based techniques in complex scenarios, HOG provides an interpretable and computationally efficient baseline for object detection in environments where edge-based representations dominate.

This study benchmarks the mentioned set of object detection architectures, namely, YOLO-v4, YOLO-v7, EfficientDet-b0, MobileNet-v1, and HOG, within the context of event-based vision. By incorporating both deep learning-based and classical approaches, we aim to assess their strengths and limitations in handling the challenges posed by event-driven data, where temporal dynamics and sparse activation patterns necessitate adaptable detection strategies.
%fill in space?

%% file: existing_approaches.tex
%\section*{Background}
\label{sec:background}
% describe event-vision
\subsection*{Existing Event-Based Vision}

%% file: proposed_method.tex
\section*{Collection Methods and Design}
\label{sec:proposed_method}
The process of collecting pixel activation statistics across images following the application of each threshold \( T_h \) is detailed in Algorithm \ref{alg:activated_pixel_detection}. Along each image column \( M_i \), the total count of activated pixels—represented as positive binary values—is computed. 

After obtaining the global activation count \( A_{\text{total}} \) from the entire image, we extract activation values specific to object regions defined by bounding boxes \( B = \{b_1, b_2, ..., b_n\} \). Each bounding box, initially given in normalized coordinates, is converted to absolute pixel indices, ensuring precise spatial segmentation.

For each bounding box \( b_i \), the subset of activated pixels within its region, denoted as \( A_{b_i} \), is computed. This provides an object-wise activation metric, which is essential for analyzing localized responses to different thresholds.

To assess background activity and non-object activation, a complementary masking approach is applied. A binary mask \( M_{\text{outside}} \) of size \( (M, N) \) is initialized with ones and subsequently updated by setting pixels within bounding boxes to zero. The remaining active pixels outside the bounding boxes are then summed to compute \( A_{\text{non-bbox}} \), representing activation in non-object regions.

Finally, the proportion of activation outside bounding boxes, \( P_{\text{non-bbox}} \), is derived as:
\begin{equation}
P_{\text{non-bbox}} = \left(\frac{A_{\text{non-bbox}}}{A_{\text{total}}}\right) \times 100
\end{equation}
This metric quantifies the relative presence of activated pixels in the background, serving as a measure of noise or irrelevant activations in the image.

\begin{algorithm}
    \caption{Detection of Activated Pixels on Multiple Objects}
    \label{alg:activated_pixel_detection}
    \begin{algorithmic}[1]
        \State \textbf{Input:} Image $I$ of size $(M,N)$, Bounding Boxes $B = \{b_1, b_2, ..., b_n\}$
        \State \textbf{Output:} Activated pixel counts for entire image, objects, and non-object areas
        \State \textbf{Step 1: Load and Preprocess Image}
        \State Convert $I$ to a binary activation map $A$ (1 for active pixels, 0 otherwise)
        \State \textbf{Step 2: Compute Global Activation}
        \State $A_{\text{total}} \gets \sum A$ \Comment{Sum of all activated pixels in the image}

        \State \textbf{Step 3: Convert Bounding Boxes to Absolute Coordinates}
        
        \ForAll{$b_i = (x_c, y_c, w, h) \in B$}
            \State Compute $x_{\min}, y_{\min}, x_{\max}, y_{\max}$ from normalized $b_i$
        \EndFor
        
        \State \textbf{Step 4: Compute Activation Within Each Bounding Box}
        
        \ForAll{$b_i \in B$}
            \State $A_{b_i} \gets \sum A[x_{\min}:x_{\max}, y_{\min}:y_{\max}]$
        \EndFor

        \State \textbf{Step 5: Compute Activation Outside Bounding Boxes}
        \State Initialize mask $M_{\text{outside}}$ of size $(M, N)$ as all ones
        
        \ForAll{$b_i \in B$}
            \State Set $M_{\text{outside}}[x_{\min}:x_{\max}, y_{\min}:y_{\max}] = 0$
        \EndFor
        
        \State $A_{\text{non-bbox}} \gets \sum (A \cdot M_{\text{outside}})$
        \State \textbf{Step 6: Compute Percentage Metrics}
        \State $P_{\text{non-bbox}} = (A_{\text{non-bbox}} / A_{\text{total}}) \times 100$ \Comment{Noise \%}
        \State \textbf{Return} $A_{\text{total}}, A_{\text{bboxes}}, A_{\text{non-bbox}}, P_{\text{non-bbox}}$
    \end{algorithmic}
\end{algorithm}

\begin{table*}[h] \scriptsize
    \vspace{-0.25cm}
    \renewcommand{\arraystretch}{1.5}
    \caption{Dataset attributes including video and image characteristics based on different threshold values.
    \label{tab:dataset_attributes}}
    \centering
    \scalebox{1.50}{ % Adjust scaling factor
        \begin{tabular}{|c|c|c|}
            \hline
            \multirow{2}{*}{Attribute} & \multirow{2}{*}{Videos} & \multirow{2}{*}{Images (each $T_h$)} \\
            & & \\
            \hline
            Number of Samples & 33 & 3,039 \\
            \hline
            Total Dataset Size & 323.5 MB (.zip) & 867.9 MB (.zip) \\
            \hline
            Length of Videos (sec) & 3.4 – 9.07 & N/A \\
            \hline
            Frame Rate (FPS) & 15 & N/A \\
            \hline
            File Format & MP4 & JPG \\
            \hline
            Average Image Resolution (AIR) & 1920 × 1080 & 416 × 416 \\
            \hline
            Aspect Ratio & 16:9 & 1:1 \\
            \hline
            Threshold Range ($T_h$) & Full Frame (0) & 4, 8, 12, 16, 20, 30, 40, 50, 60, 75 \\
            \hline
        \end{tabular}
    }
    %\vspace{-0.6cm}
\end{table*}
\subsection*{Description of Source Dataset}
\label{sec:aff_wild}
%Joey - ok
As our input source, we use the publicly available pedestrian crossing dataset; the Nanyan Technology University (NTU) Pedestrian Dataset \cite{NTU:2017},\cite{NTU:2019}. The dataset consists of 35 crossing and 35 stopping (not-crossing) scenarios where pedestrians and vehicles are present. Due to the necessity of movement within each frame to calculate event-based images, we chose not to utilize the 35 stopping videos. We also discarded two crossing videos due to a lack of objects and video length. In total, we select a subset of 33 videos from the original dataset to do our analysis. The assorted metadata regarding the NTU Pedestrian Dataset's videos and the corresponding transformed frames that had been extracted can be seen in Table \ref{tab:dataset_attributes}. The videos ranged from 3.4 to 9.1 seconds, having 51 to 136 frames, respectively. Each frame's resolution size was reduced from 1920x1080 to 416x416 for a 1:1 ratio, organized for various CNNs like Darknet and YOLO-based models. In total, the 33 videos were extracted to $\sim$3k full-frame images to be annotated, and then processed for pixel activity using the ten $T_h$ values, [4, 8, 12, 16, 20, 30, 40, 50, 60, 75]. In total the resulting amount of images is $\sim$30k, depicted in Figure \ref{fig:dataset_organization}.

% Joey - ok
\subsection*{Image Annotation Guideline}
\label{sec:img_anno}

The labeling of a dataset is a fundamental step in training successful object detection models. Each piece of trainable data is dependent on its detailed annotation and can be organized as a pair \( D(x_i, y_i) \), where \( x_i \) denotes the image, and \( y_i \) is the corresponding annotation. For this study, a team of four annotators performed the image annotations, \( y_i \), using Roboflow \cite{Dwyer_Roboflow:2022} to accurately draw the boundaries and export associated formatted files for classification and object detection. The annotators, possessing practical experience and domain-specific knowledge, ensured consistency and precision throughout the labeling process. Each image, \( x_i \), was carefully examined and bounding boxes were manually placed to accurately encapsulate the objects of interest while minimizing unnecessary background.

To maintain labeling uniformity, a structured approach was followed. Bounding boxes were annotated from the bottom-left corner \(x_{min}, y_{min}\) to the top-right corner \(x_{max}, y_{max}\) of each object \cite{Papadopoulos:2017}, according to a predefined class taxonomy (\textit{person}, \textit{vehicle}), ensuring consistent categorization across all images. In cases of partial occlusion, objects were labeled only when enough distinguishing features were visible for reliable identification. Additionally, annotation quality was reinforced through a cross-validation process, where multiple annotators reviewed and refined the labeled data to minimize errors and discrepancies. The annotation process was designed to capture the full extent of all pedestrians crossing the street as well as all vehicles in motion. Each bounding box was drawn to enclose the entire pedestrian as they moved, ensuring that variations in body posture, such as extended legs during walking, were fully accounted for. Similarly, vehicles were annotated throughout their motion, maintaining consistency across frames.

Since pedestrian movements naturally result in variations in the size of the bounding box, the annotations were dynamically adapted to reflect these changes without compromising accuracy. In cases when multiple pedestrians were walking together and their bodies overlapped, annotators carefully delineated separate bounding boxes for each person, ensuring that individual instances were preserved as accurately as possible. In such cases, the bounding boxes were adjusted to minimize occlusion effects while maintaining the integrity of the dataset.

%% file: applied_analysis.tex
% Joey - maybe add more, ok
\begin{figure*}[t!]
 \centering
    \vspace{-0.0cm}
    \includegraphics[width=0.75\textwidth]{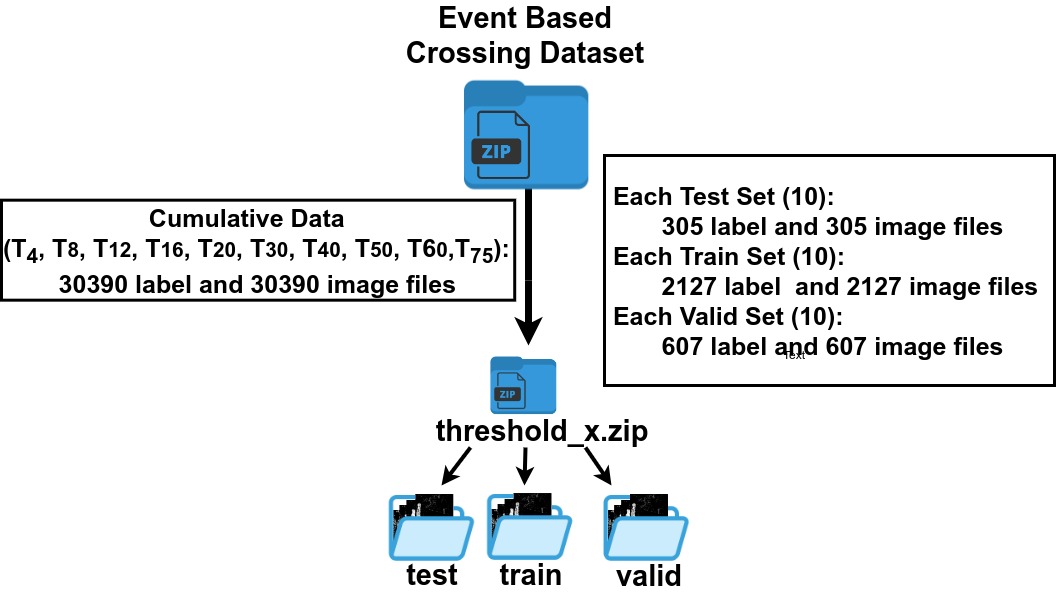}
    \caption[ ]{Dataset directories and files organization}
    \label{fig:dataset_organization}
    \vspace{-0.6cm}
\end{figure*}
\section*{Applied Analysis}
\begin{figure*}[h] 
 \centering
\includegraphics[width=0.60\textwidth]{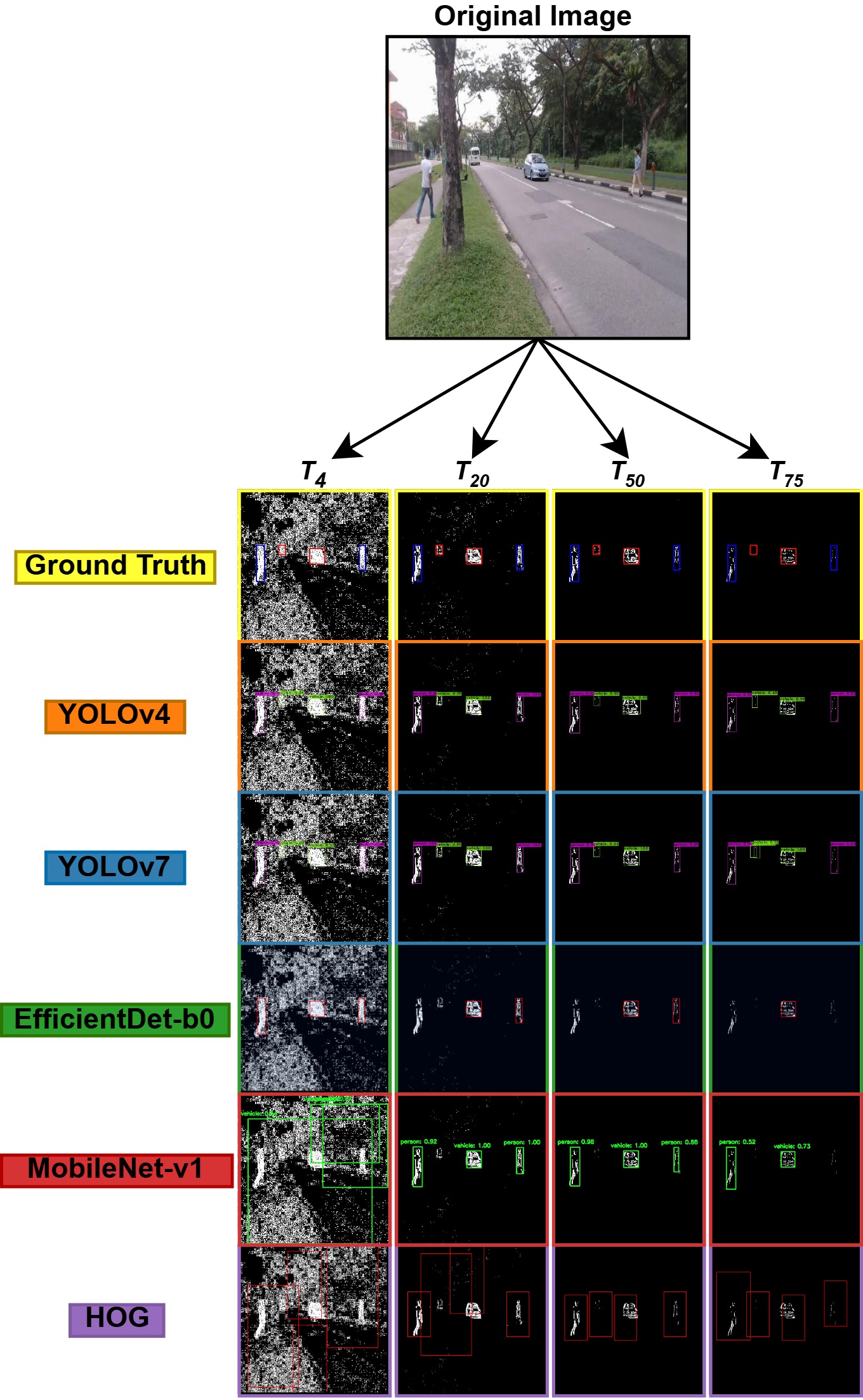}
    \caption{
        Comparison of object detection model inferences across varying threshold levels. The original image contains four objects: two pedestrians and two vehicles. The first row displays the baseline ground truth (yellow) bounding boxes for reference. As $T_h$ increases, detections become more selective, reducing false positives while incorporating less noise. } 
    \label{fig:bbox_compare}
    \vspace{-0.7cm}
\end{figure*}

Object detection plays a critical role in various computer vision applications, ranging from autonomous navigation \cite{Kitti:2012} to surveillance and industrial automation \cite{Maschler:2021}. The ability of a detection model to accurately identify and localize objects under varying event-based thresholds with high confidence is essential for ensuring reliability in practical scenarios. Figure \ref{fig:bbox_compare} provides a comparative, bounding box, analysis of multiple SOTA object detection models across various increasing $T_h$ levels, illustrating how each model adapts to variations in pixel activity and object localization selectivity. By visualizing the detection of each model, we can utilize a human-understandable reference for intuitively evaluating the most effective $T_h$ level to test model robustness and precision.
As the $T_h$ is incremented, detections become more accurate, reducing false positives and filtering out low-confidence bounding boxes. Each row represents the detections from the specified SOTA model. YOLO-v4 (orange) \cite{Bochkovskiy_yolov4:2020} and YOLO-v7 (blue) \cite{Wang_yolov7:2022} maintain consistent detections across thresholds present in the Figure ($T_4$, $T_{20}$, and $T_{50}$, $T_{75}$), demonstrating resilience to threshold variations. HOG (purple) \cite{HOG:2005}, a traditional non-neural network-based approach, while it's itersection over union (IoU) remains low, the method exhibits high sensitivity to threshold changes, drastically improving its detections from $T_4$ and $T_{20}$, moving to $T_{50}$. MobileNet-v1 (red) \cite{Howard_mobilenets:2017} also shows a notable increase in accurate detections, increasing its IoU performance when transitioning from $T_4$ to $T_{20}$, indicating improved localization accuracy as noise is reduced. Conversly, EfficientDet-b0 (green) \cite{Tan_efficientdet:2020} exhibits a decline in the number of correctly identified objects as $T_h$ increases, suggesting a loss of recall due to over-filtering of pixels. These findings highlight the importance of  selecting an appropriate $T_h$ value when deploying object detection models in event-based imaging application. A well-calibrated threshold can balance precision and recall, ensuring fewer redundant activations.

%Joey - ok
\section*{Records and Storage}
We organize the dataset into a hierarchy structure of zipped (ZIP) directories. The Event Based Crossing Dataset organization can be seen in Figure \ref{fig:dataset_organization}, where the entirety of all data is stored within a single ZIP folder containing 30390 total images and labels. These images and labels are then broken down into respective ZIP folders for each threshold ($T_h$) established in this research ($T_4$, $T_8$, $T_{12}$, $T_{16}$, $T_{20}$, $T_{30}$, $T_{40}$, $T_{50}$, $T_{60}$, $T_{75}$). Each threshold has its own directory (threshold\_4.zip, \ldots , threshold\_75.zip), housing a subset for testing, training, and validation data, having 305, 2127, and 607 images and labels, respectively, adhering to common machine learning data practices. The corresponding dataset split contains its images in the form of .jpg and its labels in .txt, similar to other event datasets~\cite{islam2024descriptorfacedetectiondataset}. The .txt label files are indexed with the classes and normalized bounding box coordiantes of a given .jpg image, formatted with the Darknet framework labeling structure.

% Joey - shorten caption
\begin{figure*}[t!]
    \centering
    \includegraphics[width=0.98\textwidth]{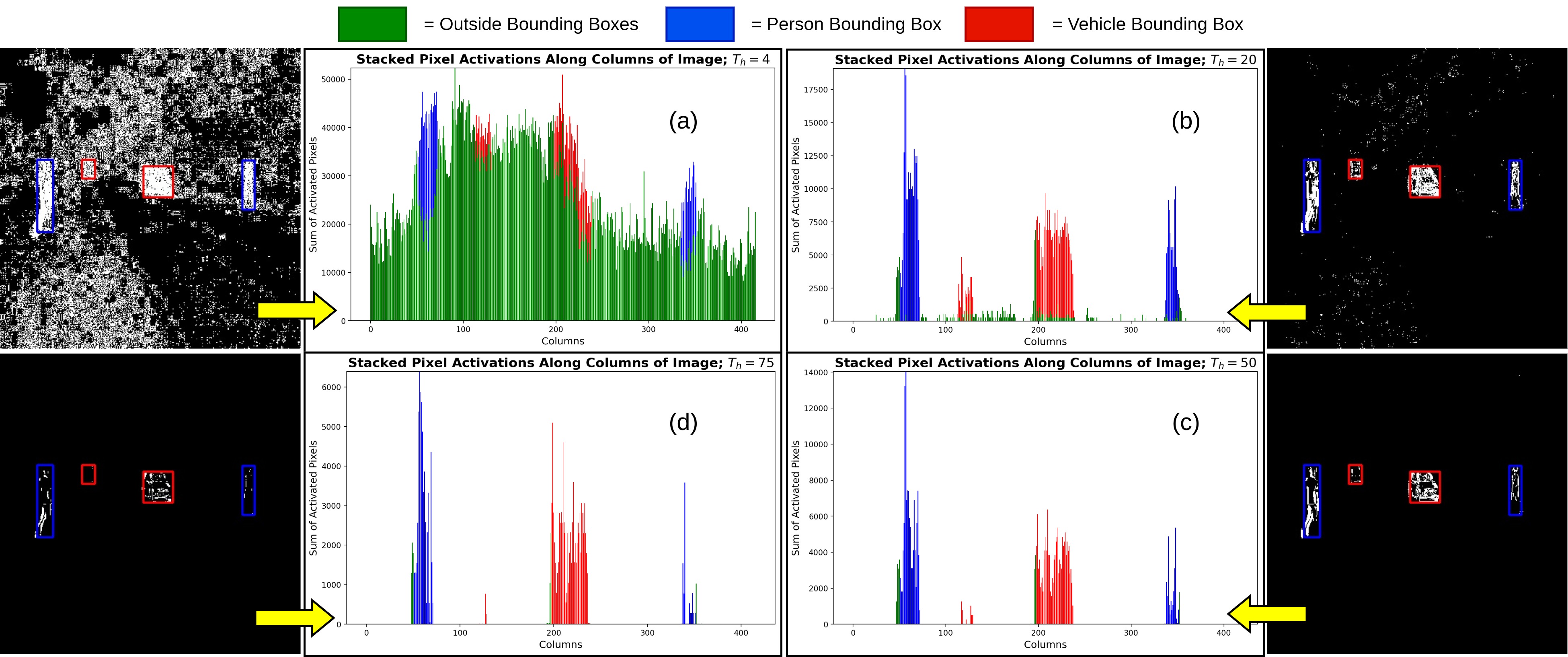}
    \caption[ ]{
        Stacked pixel activations for $T_4$ (a), $T_{20}$ (b), $T_{50}$ (c), and $T_{75}$ (d), illustrating noise suppression with increasing thresholds. At $T_4$, extensive pixel activations obscure object boundaries, with 90.09\% of activations occurring outside bounding boxes. As the threshold increases, background noise diminishes, reducing extraneous activations to 14.24\%, 6.62\%, and 5.96\% at $T_{20}$, $T_{50}$, and $T_{75}$, respectively, improving object visibility and segmentation.
        }
    \label{fig:stacked_activations}
    \vspace{-0.6cm}
\end{figure*}

%Joey - ok
\subsection*{Pixel Activity Calculation and Comparison}
When visualizing data, particularly data extracted from outdoor environments, event-based imaging presents unique challenges in pixel activation analysis. The hypersensitivity of event-based methods can capture even the slightest changes within the frame, resulting in high temporal false positives, especially when only specific actors or objects are of interest. Environmental factors such as subtle variations in lighting, shadows, or weather conditions can trigger unintended events, leading to prevalent noise and false detections \cite{Miao:2019}. These factors highlight the need for robust filtering or thresholding techniques to ensure a meaningful calculation and representation of pixel activation. 

Given the sensitivity of event-based imaging, a range of thresholds is applied to suppress unnecessary pixel activations. As shown in Figure~\ref{fig:stacked_activations}(a), the sum of activated pixels outside the bounding boxes at $T_4$ accounts for 90.09\% of all activations, significantly reducing detection efficacy for targeted objects. Increasing the threshold reduces the proportion of noisy pixels by up to $15\times$, from $T_4$ to $T_{75}$, thereby improving the signal-to-noise ratio and enhancing object segmentation. This thresholding process mitigates background noise but also affects object-level feature retention. Lower thresholds preserve a greater number of activated pixels but compromise spatial integrity due to excessive noise, making object boundaries less distinct. Higher thresholds reduce background activations, sharpening object contours and improving localization but at the cost of finer details, particularly for smaller objects. This trade-off between noise suppression and object preservation underscores the necessity of selecting an optimal $T_h$ based on specific application requirements. The variation in pixel activation across thresholds reveals the extent to which noise influences object visibility. At $T_4$, widespread extraneous activations overlap with designated bounding boxes, obscuring object details. As the threshold increases, background noise diminishes, leading to a more precise representation of detected objects. Quantitatively, the proportion of activated pixels outside bounding boxes decreases from 90.09\% at $T_4$ to 14.24\% at $T_{20}$, 6.62\% at $T_{50}$, and 5.96\% at $T_{75}$. This progressive reduction in non-relevant activations enhances object interpretability and improves event-based detection accuracy by filtering out environmental disturbances while retaining critical object features.

To further illustrate the distinction between noisy and structured activations within different environments, we compute the pixel activation distributions for two sample images: one from this dataset, representing a variable outdoor environment, EBCD, and another from a controlled indoor environment of the SEFD~\cite{SEFD:2024}. As presented in Figure~\ref{fig:activity_compare}, both images are analyzed at $T_4$—the lowest threshold considered—to emphasize the impact of environmental factors on pixel activation patterns. Although the detection tasks differ, with pedestrian detection in EBCD, and facial recognition in SEFD, this comparison strictly evaluates pixel activity independent of object type. The outdoor sample from EBCD demonstrates significantly higher activation variance, with more dispersed and irregular pixel activations. Environmental factors such as lighting variations, background movement, and external stimuli contribute to a noisier activation map, making object localization more challenging. In contrast, the indoor sample from SEFD presents a more structured activation pattern, with activations concentrated around desired object features. The controlled lighting and reduced external interference in SEFD contribute to a more predictable and stable activation distribution. The contrast between these two environments is quantitatively evident, as the EBCD image exhibits a $5.2\times$ increase in average active pixels per column compared to SEFD. 

% These findings underscore the necessity of thresholding and adaptive filtering in event-based vision, particularly in highly dynamic settings where noise management is critical.

The interplay between threshold selection, environmental conditions, and activation density underscores the challenges of event-based imaging in uncontrolled settings. Whereas many DVS datasets that employ a fixed activation threshold for all images \cite{Orchard:2015}, \cite{Li:2017CIFAR10DVS}, \cite{Kim:2021NImageNet}, our approach demonstrates the benefits of applying a range of thresholds to adapt to varying scene conditions. A single predefined threshold may fail to generalize across diverse environments, as lower thresholds retain excessive noise while higher thresholds risk omitting finer object details. 
%removed for space
%By providing a threshold range, we allow for a more flexible evaluation of pixel activations, enabling a balance between noise suppression and object retention based on specific detection requirements given in pedestrian detection circumstances.

%% file: analysis.tex
% Joey - ok
\section*{Insights and Notes}
\label{sec:analysis}

\subsection*{Benchmarking with State-of-the-Art Architectures and Techniques}

The evaluation of this dataset employs a diverse set of object detection models, aforementioned: YOLO-v4, YOLO-v7, EfficientDet-b0, MobileNet-v1, and HOG. All analysis was caputed on an Intel Xeon(R) 20-core CPU with 32 GB RAM and a 8 GB NVIDIA Quadro P4000 Graphics card running Ubuntu 20.04.6 LTS. Each model is assessed across the full range of activation thresholds presented in Table~\ref{tab:benchmarks}: $T_4$, $T_8$, $T_{12}$, $T_{16}$, $T_{20}$, $T_{30}$, $T_{40}$, $T_{50}$, $T_{60}$, and $T_{75}$. This range allows for a comprehensive analysis of detection performance under varying levels of activation filtering, a critical aspect in event-based vision where noise suppression plays a key role in ensuring reliable object localization~\cite{Miao:2019}.

\begin{figure*}[h!]
\centering 
%\vspace{-0.1cm}
\includegraphics[width=1.00\textwidth]{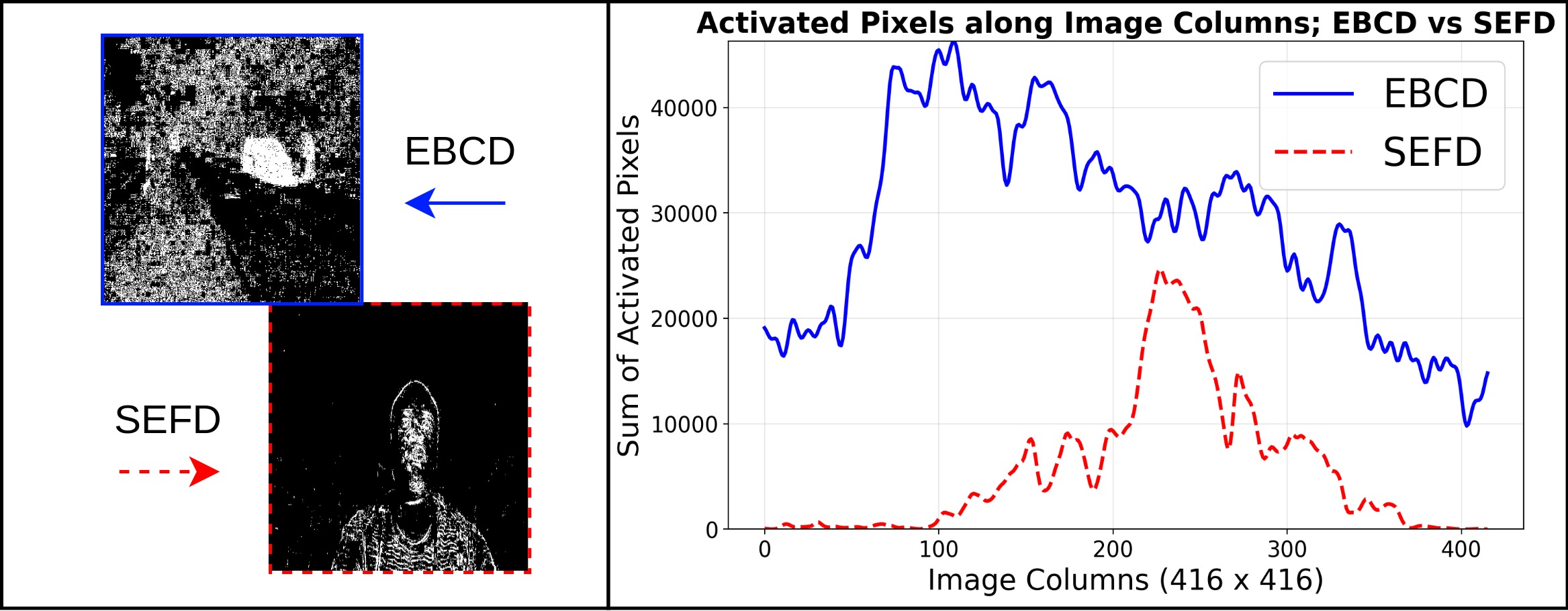}
    \caption[]{
        Pixel activation distributions for an outdoor event-based image (EBCD, blue) and an indoor event-based image (SEFD, red) at $T_4$. The outdoor image exhibits significantly higher and more dispersed activations due to environmental noise, whereas the indoor image presents a more localized and controlled activation pattern.}
    \label{fig:activity_compare}
    \vspace{-0.6cm}
\end{figure*}

\begin{figure*}[t!] 
\centering  
\includegraphics[width=1.00\textwidth]{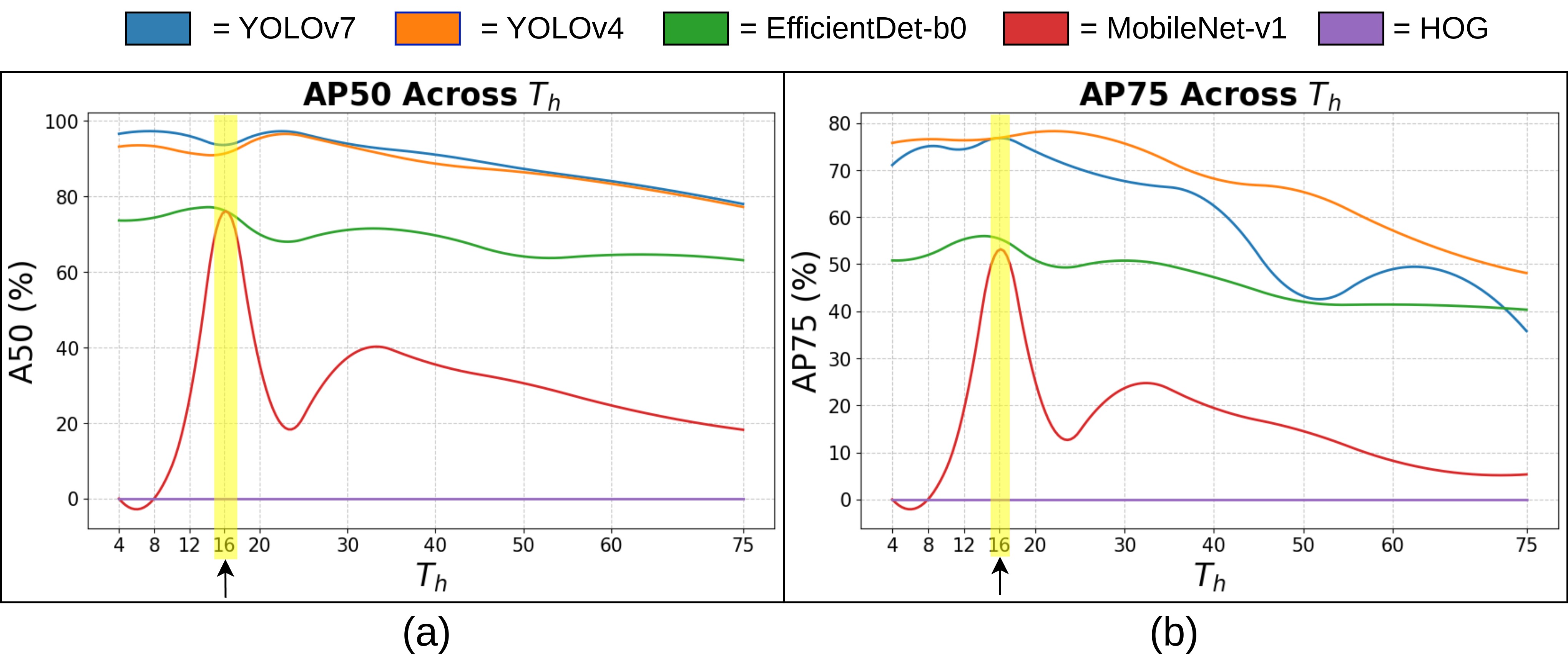}
    \caption[ ]{
        The graphs present the AP50 (a) and AP75 (b) across different decision thresholds object detection models: YOLO-v4, YOLO-v7, EfficientDet-b0, MobileNet-v1, and HOG. The yellow vertical highlight indicates the optimal threshold ($T_{16}$) that yields the best average performance across models.
        } 
    \label{fig:ap50_ap75_compare}
\end{figure*}
%\vspace{-0.4cm}

In order to determine the best overall threshold, we established in Figure~\ref{fig:ap50_ap75_compare}, the impact of different thresholds on average precision at IoU 50\% (AP50) and average precision at IoU 75 (AP75), highlighting an optimal threshold range for maximizing detection performance. The results indicate that $T_{16}$ provides the best overall balance between noise suppression and object retention. As the threshold increases beyond this point, both AP50 and AP75 exhibit a decline for certain models, particularly EfficientDet-b0 and MobileNet-v1. This decline can be attributed to the trade-off between filtering noisy activations and retaining sufficient object features for accurate classification and localization~\cite{Gallego:2022}.

Additionally, the HOG-based detector consistently fails to achieve IoU values above 50\% or 75\%, resulting in AP50 and AP75 scores of zero across all thresholds. This outcome is reflected both numerically in Table~\ref{tab:benchmarks} and visually in Figure~\ref{fig:bbox_compare}. The inability of HOG to meet these thresholds suggests a fundamental limitation of traditional gradient-based descriptors in handling the spatial and temporal sparsity inherent in event-based imaging. However, HOG continues to attempt its localization and generates bounding boxes around the correct objects, albeit with very low IoU, thus not reaching the threshold of calculating a correct detection. Deep learning models, by contrast, leverage hierarchical feature extraction, enabling them to adapt to varying activation densities and complex spatial representations~\cite{Gallego:2022}.
\begin{figure*}[t!]
 \centering
    \includegraphics[width=1.0\textwidth]{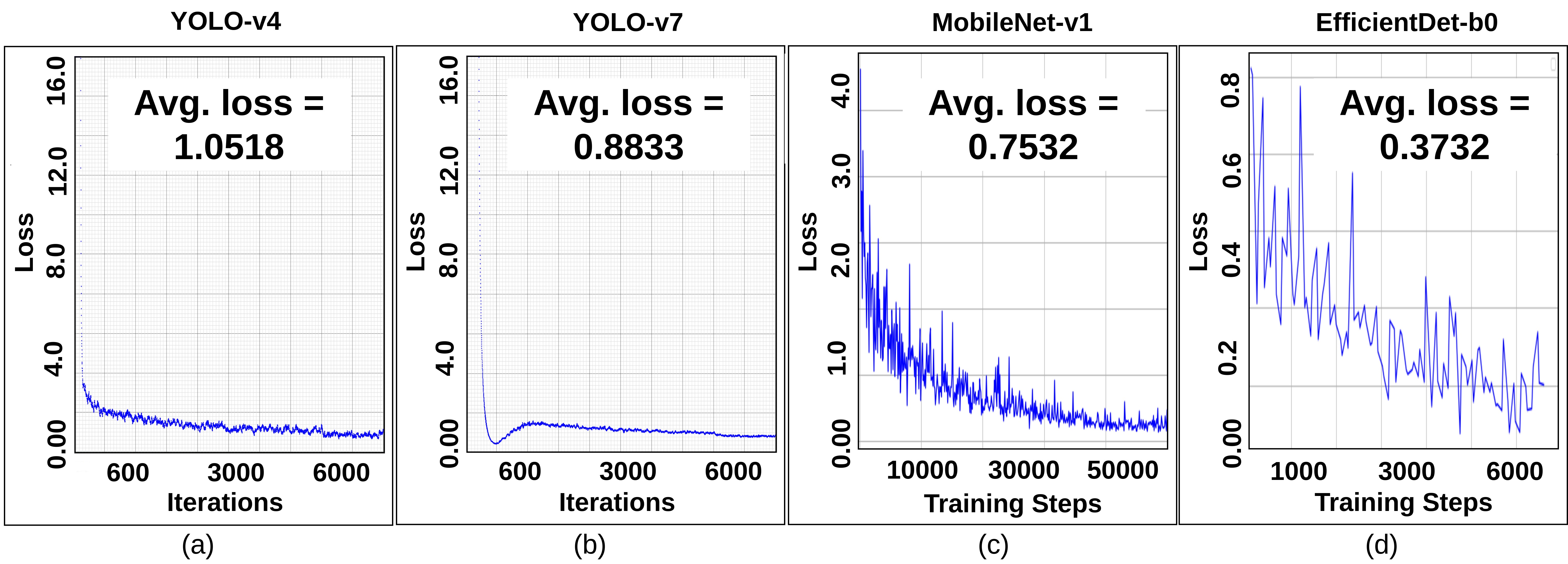}
    \caption[ ]{
        Loss curves for YOLO-v4 (a), YOLO-v7 (b), MobileNet-v1 (c), and EfficientDet-b0 (d), illustrating their training loss over time.
        } 
    \label{fig:loss_curves}
\end{figure*}
%\vspace{-0.9cm}

To further analyze model performance, Figure~\ref{fig:loss_curves} presents the training loss curves for the deep learning-based detectors on the calculated best $T_h$ value; $T_{16}$. These curves provide insight into model convergence, optimization stability, and potential overfitting or underfitting issues. A well-behaved loss curve should exhibit a steady decline and eventual stabilization, indicating effective learning and generalization. However, in event-based vision, training loss can be influenced by the inherent sparsity and non-redundant nature of event data, which may require specialized augmentation or domain-adaptive learning strategies to achieve optimal performance~\cite{Zhu:2019}. Before examining, it is essential to distinguish between iterations and training steps, as different deep learning frameworks define these terms in distinct ways. In Darknet-based models (YOLO-v4 and YOLO-v7), training progresses in iterations, where each iteration corresponds to a forward and backward pass over a single batch of data. Conversely, TensorFlow-based models (MobileNet-v1 and EfficientDet-b0) use training steps, which represent updates to the model’s parameters after processing a batch. While the concepts are similar, numerical differences arise due to variations in how frameworks handle batch accumulation and gradient updates.

YOLO-v4 achieves an average loss of 1.0518, demonstrating a steady decrease over 6000 iterations, while YOLO-v7 reaches a lower loss of 0.8833, following a similar convergence pattern. MobileNet-v1, trained over a significantly larger number of steps (50,000), stabilizes at 0.7532 but exhibits greater fluctuations in the early stages, reflecting a more variable optimization process. EfficientDet-b0 maintains the lowest loss of 0.3732 but shows considerable variance throughout training, suggesting a higher sensitivity to weight initialization and hyperparameter tuning. These loss curves highlight key differences in optimization behavior, convergence speed, and overall stability across models.

%% file: main.bbl
\begin{thebibliography}{10}
\providecommand{\url}[1]{#1}
\csname url@samestyle\endcsname
\providecommand{\newblock}{\relax}
\providecommand{\bibinfo}[2]{#2}
\providecommand{\BIBentrySTDinterwordspacing}{\spaceskip=0pt\relax}
\providecommand{\BIBentryALTinterwordstretchfactor}{4}
\providecommand{\BIBentryALTinterwordspacing}{\spaceskip=\fontdimen2\font plus
\BIBentryALTinterwordstretchfactor\fontdimen3\font minus \fontdimen4\font\relax}
\providecommand{\BIBforeignlanguage}[2]{{%
\expandafter\ifx\csname l@#1\endcsname\relax
\typeout{** WARNING: IEEEtran.bst: No hyphenation pattern has been}%
\typeout{** loaded for the language `#1'. Using the pattern for}%
\typeout{** the default language instead.}%
\else
\language=\csname l@#1\endcsname
\fi
#2}}
\providecommand{\BIBdecl}{\relax}
\BIBdecl

\bibitem{Lichtsteiner:2008}
P.~Lichtsteiner, C.~Posch, and T.~Delbruck, ``A 128$\times$ 128 120 db 15 $\mu$s latency asynchronous temporal contrast vision sensor,'' \emph{IEEE Journal of Solid-State Circuits}, vol.~43, no.~2, pp. 566--576, 2008.

\bibitem{EBV:A_Suvey}
G.~Gallego, T.~Delbrück, G.~Orchard, C.~Bartolozzi, B.~Taba, A.~Censi, S.~Leutenegger, A.~J. Davison, J.~Conradt, K.~Daniilidis, and D.~Scaramuzza, ``Event-based vision: A survey,'' \emph{IEEE Transactions on Pattern Analysis and Machine Intelligence}, vol.~44, no.~1, pp. 154--180, 2022.

\bibitem{islam2023exploring}
R.~Islam, P.~Majurski, J.~Kwon, and S.~R. S.~K. Tummala, ``{Exploring high-level neural networks architectures for efficient spiking neural networks implementation},'' in \emph{2023 3rd International Conference on Robotics, Electrical and Signal Processing Techniques (ICREST)}.\hskip 1em plus 0.5em minus 0.4em\relax IEEE, 2023, pp. 212--216.

\bibitem{caporale2008spike}
N.~Caporale and Y.~Dan, ``{Spike timing--dependent plasticity: a Hebbian learning rule},'' \emph{Annu. Rev. Neurosci.}, vol.~31, no.~1, pp. 25--46, 2008.

\bibitem{islam2024benchmarking}
R.~Islam, P.~Majurski, J.~Kwon, A.~Sharma, and S.~R. S.~K. Tummala, ``Benchmarking artificial neural network architectures for high-performance spiking neural networks,'' \emph{Sensors}, vol.~24, no.~4, p. 1329, 2024.

\bibitem{Amir_dvs:2017}
A.~Amir, B.~Taba, D.~Berg, T.~Melano, J.~McKinstry, C.~Di~Nolfo, T.~Nayak, A.~Andreopoulos, G.~Garreau, M.~Mendoza, J.~Kusnitz, M.~Debole, S.~Esser, T.~Delbruck, M.~Flickner, and D.~Modha, ``A low power, fully event-based gesture recognition system,'' in \emph{IEEE Conference on Computer Vision and Pattern Recognition (CVPR)}, 2017, pp. 7388--7397.

\bibitem{Gallego:2022}
G.~Gallego, T.~Delbrück, G.~Orchard, C.~Bartolozzi, B.~Taba, A.~Censi, S.~Leutenegger, A.~J. Davison, J.~Conradt, K.~Daniilidis, and D.~Scaramuzza, ``Event-based vision: A survey,'' \emph{IEEE Transactions on Pattern Analysis and Machine Intelligence}, vol.~44, no.~1, pp. 154--180, 2022.

\bibitem{Rebecq:2019}
H.~Rebecq, R.~Ranftl, V.~Koltun, and D.~Scaramuzza, ``Events-to-video: Bringing modern computer vision to event cameras,'' in \emph{Proceedings of the IEEE/CVF Conference on Computer Vision and Pattern Recognition}.\hskip 1em plus 0.5em minus 0.4em\relax IEEE, 2019, pp. 3857--3866.

\bibitem{Islam_asicon:2011}
R.~Islam, S.~Esmaeili, and T.~Islam, ``{A high performance clock precharge SEU hardened flip-flop},'' in \emph{IEEE International Conference on ASIC}, 2011, pp. 574--577.

\bibitem{mitra2007built}
S.~Mitra, M.~Zhang, N.~Seifert, T.~Mak, and K.~S. Kim, ``{Built-in soft error resilience for robust system design},'' in \emph{2007 IEEE International Conference on Integrated Circuit Design and Technology}.\hskip 1em plus 0.5em minus 0.4em\relax IEEE, 2007, pp. 1--6.

\bibitem{Islam_isqed:2012}
R.~Islam, ``{A highly reliable SEU hardened latch and high performance SEU hardened flip-flop},'' in \emph{Thirteenth International Symposium on Quality Electronic Design (ISQED)}, 2012, pp. 347--352.

\bibitem{islam2018low}
------, ``{Low-power resonant clocking using soft error robust energy recovery flip-flops},'' \emph{Journal of Electronic Testing}, vol.~34, no.~4, pp. 471--485, 2018.

\bibitem{Miao:2019}
\BIBentryALTinterwordspacing
S.~Miao, G.~Chen, X.~Ning, Y.~Zi, K.~Ren, Z.~Bing, and A.~Knoll, ``Neuromorphic vision datasets for pedestrian detection, action recognition, and fall detection,'' \emph{Frontiers in Neurorobotics}, vol.~13, 2019. [Online]. Available: \url{https://www.frontiersin.org/journals/neurorobotics/articles/10.3389/fnbot.2019.00038}
\BIBentrySTDinterwordspacing

\bibitem{1MEGAPIXEL}
\BIBentryALTinterwordspacing
E.~Perot, P.~de~Tournemire, D.~Nitti, J.~Masci, and A.~Sironi, ``Learning to detect objects with a 1 megapixel event camera,'' in \emph{Advances in Neural Information Processing Systems}, vol.~33, 2020, pp. 16\,639--16\,652. [Online]. Available: \url{https://papers.nips.cc/paper/2020/file/c213877427b46fa96cff6c39e837ccee-Paper.pdf}
\BIBentrySTDinterwordspacing

\bibitem{EventBasedPedestrain}
\BIBentryALTinterwordspacing
Y.~Yao, Z.~Huang, Y.~Chen, Y.~Zhang, and Y.~Wang, ``Event-based pedestrian detection using dynamic vision sensors,'' \emph{Electronics}, vol.~10, no.~8, p. 888, 2021. [Online]. Available: \url{https://www.mdpi.com/2079-9292/10/8/888}
\BIBentrySTDinterwordspacing

\bibitem{EventBasedPedestrian_IEEE}
\BIBentryALTinterwordspacing
------, ``Event-based pedestrian detection using dynamic vision sensors,'' in \emph{Proceedings of the IEEE Conference on Computer Vision and Pattern Recognition}, 2021, pp. 1110--1119. [Online]. Available: \url{https://ieeexplore.ieee.org/document/5206631}
\BIBentrySTDinterwordspacing

\bibitem{Chen:2018}
G.~Chen, H.~Cao, M.~Aafaque, J.~Chen, and C.~Ye, ``Neuromorphic vision-based multivehicle detection and tracking for intelligent transportation system,'' \emph{Journal of Advanced Transportation}, vol. 2018, 2018.

\bibitem{Boretti:2023}
C.~Boretti, P.~Bich, F.~Pareschi, L.~Prono, R.~Rovatti, and G.~Setti, ``Pedro: An event-based dataset for person detection in robotics,'' in \emph{Proceedings of the IEEE/CVF Conference on Computer Vision and Pattern Recognition Workshops}.\hskip 1em plus 0.5em minus 0.4em\relax IEEE, 2023, pp. 1--8.

\bibitem{SEFD:2024}
R.~Islam, S.~R. S.~K. Tummala, J.~Mulé, R.~Kankipati, S.~Jalapally, D.~Challagundla, C.~Howard, and R.~Robucci, ``Descriptor: Smart event face dataset (sefd),'' \emph{IEEE Data Descriptions}, vol.~1, pp. 33--40, 2024.

\bibitem{Afshar:2020}
\BIBentryALTinterwordspacing
S.~Afshar, N.~Ralph, Y.~Xu, J.~Tapson, A.~van Schaik, and G.~Cohen, ``Event-based feature extraction using adaptive selection thresholds,'' \emph{Sensors}, vol.~20, no.~6, p. 1600, 2020. [Online]. Available: \url{https://www.mdpi.com/1424-8220/20/6/1600}
\BIBentrySTDinterwordspacing

\bibitem{Chakravarthi_survey:2024}
\BIBentryALTinterwordspacing
B.~Chakravarthi, A.~A. Verma, K.~Daniilidis, C.~Fermuller, and Y.~Yang, ``Recent event camera innovations: A survey,'' 2024. [Online]. Available: \url{https://arxiv.org/abs/2408.13627}
\BIBentrySTDinterwordspacing

\bibitem{islam2018negative}
R.~Islam, ``{Negative capacitance clock distribution},'' \emph{IEEE Transactions on Emerging Topics in Computing}, vol.~9, no.~1, pp. 547--553, 2018.

\bibitem{Zhang_event:2024}
A.~Zhang, J.~Pang, H.~Wu, Q.~Tan, Z.~Zheng, L.~Xu, J.~Tang, and G.~Niu, ``Event-based x-ray imager with ghosting-free scintillator film,'' \emph{Optica}, vol.~11, no.~5, pp. 606--611, 2024.

\bibitem{Guo_eventlfm:2024}
R.~Guo, Q.~Yang, A.~S. Chang, G.~Hu, J.~Greene, C.~V. Gabel, S.~You, and L.~Tian, ``Eventlfm: event camera integrated fourier light field microscopy for ultrafast 3d imaging,'' \emph{Light: Science \& Applications}, vol.~13, no.~1, p. 144, 2024.

\bibitem{Dong_sevar:2024}
Y.~Dong, Z.~Chen, X.~He, L.~Li, Z.~Shu, Y.~Cao, J.~Feng, S.~Liu, C.~Li, and J.~Wang, ``Sevar: a stereo event camera dataset for virtual and augmented reality,'' \emph{Frontiers of Information Technology \& Electronic Engineering}, vol.~25, no.~5, pp. 755--762, 2024.

\bibitem{Gokarn_poster:2024}
I.~Gokarn and A.~Misra, ``Poster: Profiling event vision processing on edge devices,'' in \emph{Proceedings of the 22nd Annual International Conference on Mobile Systems, Applications and Services}, 2024, pp. 672--673.

\bibitem{Islam_dcmcs:2018}
R.~Islam, H.~A. Fahmy, P.~Y. Lin, and M.~R. Guthaus, ``{DCMCS: Highly Robust Low-Power Differential Current-Mode Clocking and Synthesis},'' \emph{IEEE Transactions on Very Large Scale Integration (VLSI) Systems}, vol.~26, no.~10, pp. 2108--2117, 2018.

\bibitem{islam2011high}
R.~Islam, ``{High-speed energy-efficient soft error tolerant flip-flops},'' Ph.D. dissertation, Concordia University, 2011.

\bibitem{guthaus2017current}
M.~Guthaus and R.~Islam, ``{Current-mode clock distribution},'' Oct.~10 2017, uS Patent 9,787,293.

\bibitem{perez2024general}
D.~P{\'e}rez-L{\'o}pez, A.~Gutierrez, D.~S{\'a}nchez, A.~L{\'o}pez-Hern{\'a}ndez, M.~Gutierrez, E.~S{\'a}nchez-Gom{\'a}riz, J.~Fern{\'a}ndez, A.~Cruz, A.~Quir{\'o}s, Z.~Xie \emph{et~al.}, ``{General-purpose programmable photonic processor for advanced radiofrequency applications},'' \emph{Nature Communications}, vol.~15, no.~1, p. 1563, 2024.

\bibitem{Islam_diff:2015}
R.~Islam, H.~Fahmy, P.-Y. Lin, and M.~R. Guthaus, ``{Differential current-mode clock distribution},'' in \emph{2015 IEEE 58th International Midwest Symposium on Circuits and Systems (MWSCAS)}, 2015, pp. 1--4.

\bibitem{challagundla2022power}
D.~Challagundla, M.~Galib, I.~Bezzam, and R.~Islam, ``{Power and skew reduction using resonant energy recycling in 14-nm FinFET clocks},'' in \emph{2022 IEEE International Symposium on Circuits and Systems (ISCAS)}.\hskip 1em plus 0.5em minus 0.4em\relax IEEE, 2022, pp. 268--272.

\bibitem{busia2024tiny}
P.~Busia, M.~A. Scrugli, V.~J.-B. Jung, L.~Benini, and P.~Meloni, ``{A tiny transformer for low-power arrhythmia classification on microcontrollers},'' \emph{IEEE Transactions on Biomedical Circuits and Systems}, 2024.

\bibitem{Orchard:2015}
\BIBentryALTinterwordspacing
G.~Orchard, A.~Jayawant, G.~K. Cohen, and N.~Thakor, ``Converting static image datasets to spiking neuromorphic datasets using saccades,'' \emph{Frontiers in Neuroscience}, vol.~9, 2015. [Online]. Available: \url{https://www.frontiersin.org/articles/10.3389/fnins.2015.00437}
\BIBentrySTDinterwordspacing

\bibitem{Serrano-gotarredona_linares-barranco:2015}
T.~Serrano-Gotarredona and B.~Linares-Barranco, ``Poker-{DVS} and {MNIST-DVS}. their history, how they were made, and other details,'' \emph{Frontiers in Neuroscience}, vol.~9, Dec 2015.

\bibitem{Redmon_yolo:2016}
J.~Redmon, S.~Divvala, R.~Girshick, and A.~Farhadi, ``You only look once: Unified, real-time object detection,'' in \emph{Proceedings of the IEEE conference on computer vision and pattern recognition}, 2016, pp. 779--788.

\bibitem{Bochkovskiy_yolov4:2020}
A.~Bochkovskiy, C.-Y. Wang, and H.-Y.~M. Liao, ``Yolov4: Optimal speed and accuracy of object detection,'' \emph{arXiv preprint arXiv:2004.10934}, pp. 1--17, 2020.

\bibitem{Wang_yolov7:2022}
C.-Y. Wang, A.~Bochkovskiy, and H.-Y.~M. Liao, ``Yolov7: Trainable bag-of-freebies sets new state-of-the-art for real-time object detectors,'' in \emph{Proceedings of the IEEE/CVF conference on computer vision and pattern recognition}, 2023, pp. 7464--7475.

\bibitem{Tan_efficientdet:2020}
M.~Tan, R.~Pang, and Q.~V. Le, ``Efficientdet: Scalable and efficient object detection,'' in \emph{Proceedings of the IEEE/CVF conference on computer vision and pattern recognition}, 2020, pp. 10\,781--10\,790.

\bibitem{Howard_mobilenets:2017}
A.~G. Howard, ``Mobilenets: Efficient convolutional neural networks for mobile vision applications,'' \emph{arXiv preprint arXiv:1704.04861}, pp. 1--9, 2017.

\bibitem{HOG:2005}
N.~Dalal and B.~Triggs, ``Histograms of oriented gradients for human detection,'' in \emph{2005 IEEE Computer Society Conference on Computer Vision and Pattern Recognition (CVPR'05)}, vol.~1, 2005, pp. 886--893 vol. 1.

\bibitem{NTU:2017}
S.~Neogi, M.~Hoy, W.~Chaoqun, and J.~Dauwels, ``Context based pedestrian intention prediction using factored latent dynamic conditional random fields,'' in \emph{2017 IEEE Symposium Series on Computational Intelligence (SSCI)}, 2017, pp. 1--8.

\bibitem{NTU:2019}
\BIBentryALTinterwordspacing
S.~Neogi, M.~Hoy, K.~Dang, H.~Yu, and J.~Dauwels, ``Context model for pedestrian intention prediction using factored latent-dynamic conditional random fields,'' \emph{IEEE Transactions on Intelligent Transportation Systems}, vol.~22, pp. 6821--6832, 2019. [Online]. Available: \url{https://api.semanticscholar.org/CorpusID:198968363}
\BIBentrySTDinterwordspacing

\bibitem{Dwyer_Roboflow:2022}
B.~Dwyer, J.~Nelson, J.~Solawetz, and et~al., ``{Roboflow} (version 1.0) [software],'' \url{https://roboflow.com}, 2022, accessed: November, 2023.

\bibitem{Papadopoulos:2017}
\BIBentryALTinterwordspacing
D.~P. Papadopoulos, J.~R.~R. Uijlings, F.~Keller, and V.~Ferrari, ``Extreme clicking for efficient object annotation,'' in \emph{Proceedings of the IEEE International Conference on Computer Vision (ICCV)}, 2017, pp. 4940--4949. [Online]. Available: \url{https://openaccess.thecvf.com/content_ICCV_2017/papers/Papadopoulos_Extreme_Clicking_for_ICCV_2017_paper.pdf}
\BIBentrySTDinterwordspacing

\bibitem{Kitti:2012}
A.~Geiger, P.~Lenz, and R.~Urtasun, ``Are we ready for autonomous driving? the kitti vision benchmark suite,'' in \emph{2012 IEEE Conference on Computer Vision and Pattern Recognition}, 2012, pp. 3354--3361.

\bibitem{Maschler:2021}
B.~Maschler and M.~Weyrich, ``Deep transfer learning for industrial automation: A review and discussion of new techniques for data-driven machine learning,'' \emph{IEEE Industrial Electronics Magazine}, vol.~15, no.~2, pp. 65--75, 2021.

\bibitem{islam2024descriptorfacedetectiondataset}
\BIBentryALTinterwordspacing
R.~Islam, S.~R. S.~K. Tummala, J.~Mulé, R.~Kankipati, S.~Jalapally, D.~Challagundla, C.~Howard, and R.~Robucci, ``{Descriptor: Face Detection Dataset for Programmable Threshold-Based Sparse-Vision},'' 2024. [Online]. Available: \url{https://arxiv.org/abs/2410.00368}
\BIBentrySTDinterwordspacing

\bibitem{Li:2017CIFAR10DVS}
\BIBentryALTinterwordspacing
H.~Li, H.~Liu, and X.~Ji, ``Cifar10-dvs: An event-stream dataset for object classification,'' \emph{Frontiers in Neuroscience}, vol.~11, p. 309, 2017. [Online]. Available: \url{https://www.frontiersin.org/articles/10.3389/fnins.2017.00309/full}
\BIBentrySTDinterwordspacing

\bibitem{Kim:2021NImageNet}
\BIBentryALTinterwordspacing
H.~Kim, D.~Gehrig, M.~Gehrig, and D.~Scaramuzza, ``N-imagenet: Towards robust, fine-grained object recognition with event cameras,'' \emph{arXiv preprint arXiv:2112.01041}, 2021. [Online]. Available: \url{https://arxiv.org/abs/2112.01041}
\BIBentrySTDinterwordspacing

\bibitem{Zhu:2019}
\BIBentryALTinterwordspacing
A.~Z. Zhu, L.~Yuan, K.~Chaney, and K.~Daniilidis, ``Unsupervised event-based learning of optical flow, depth, and egomotion,'' in \emph{Proceedings of the IEEE Conference on Computer Vision and Pattern Recognition (CVPR)}, 2019, pp. 989--997. [Online]. Available: \url{https://openaccess.thecvf.com/content_CVPR_2019/html/Zhu_Unsupervised_Event-Based_Learning_of_Optical_Flow_Depth_and_Egomotion_CVPR_2019_paper.html}
\BIBentrySTDinterwordspacing

\end{thebibliography}
